\documentclass[10pt,twocolumn,letterpaper]{article}

\usepackage[pagenumbers]{cvpr} 

\usepackage{graphicx}
\usepackage{amsmath}
\usepackage{amssymb}
\usepackage{booktabs}
\usepackage{verbatim}
\usepackage{multirow}
\usepackage{makecell}
\usepackage{cuted}

\usepackage[pagebackref,breaklinks,colorlinks]{hyperref}

\usepackage[capitalize]{cleveref}
\crefname{section}{Sec.}{Secs.}
\Crefname{section}{Section}{Sections}
\Crefname{table}{Table}{Tables}
\crefname{table}{Tab.}{Tabs.}

\newcommand{\squishlist}{
	\begin{list}{$\bullet$}
		{ \setlength{\itemsep}{0pt}
			\setlength{\parsep}{1pt}
			\setlength{\topsep}{1pt}
			\setlength{\partopsep}{0pt}
			\setlength{\leftmargin}{1.5em}
			\setlength{\labelwidth}{1em}
			\setlength{\labelsep}{0.5em} } }
\newcommand{\squishend}{\end{list} 
}

\def\cvprPaperID{9458} 
\def\confName{CVPR}
\def\confYear{2022}

\title{3D Photo Stylization: \\Learning to Generate Stylized Novel Views from a Single Image\vspace{-0.8em}}

\author{
Fangzhou Mu$^1$\thanks{}\quad
Jian Wang$^2$\thanks{}\quad
Yicheng Wu$^2$\footnotemark[2]\quad
Yin Li$^1$\footnotemark[2]
\\
$^1$University of Wisconsin-Madison\quad
$^2$Snap Research
\\
$^1${\tt\small \{fmu2, yin.li\}@wisc.edu}\quad
$^2${\tt\small \{jwang4, yicheng.wu\}@snap.com}
}

\begin{document}
\twocolumn[{
    \renewcommand\twocolumn[1][]{#1}
    \maketitle
    \vspace{-3.6em}
    \begin{center}
        \centering
        \captionsetup{type=figure}
        \includegraphics[width=.86\textwidth]{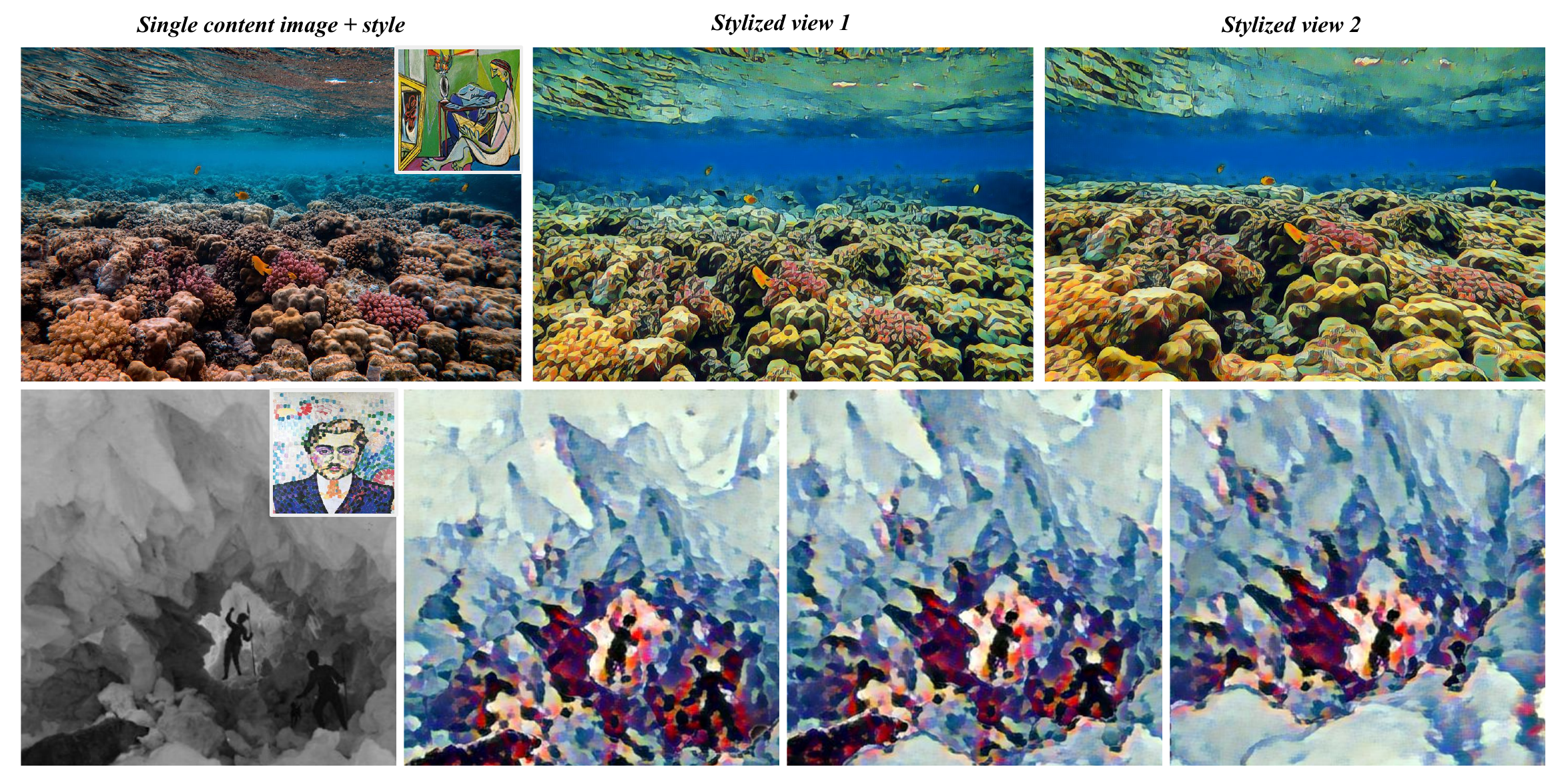}\vspace{-1.25em}
        \captionof{figure}{{\bf 3D photo stylization. } Given a {\it single} content image, our method synthesizes novel views of the scene in an arbitrary style. In doing so, our method delivers immersive viewing experience of a memorable moment within existing photos.
        }\vspace{-0.5em}
        \label{fig:teaser}
    \end{center}
}]
{
  \renewcommand{\thefootnote}
    {\fnsymbol{footnote}}
  \footnotetext[1]{Work partially done when Fangzhou was an intern at Snap Research}
  \footnotetext[2]{co-corresponding authors}
}



\begin{abstract}\vspace{-0.75em}


Visual content creation has spurred a soaring interest given its applications in mobile photography and AR / VR. Style transfer and single-image 3D photography as two representative tasks have so far evolved independently. In this paper, we make a connection between the two, and address the challenging task of 3D photo stylization --- generating stylized novel views from a single image given an arbitrary style. 
Our key intuition is that style transfer and view synthesis have to be jointly modeled for this task. To this end, we propose a deep model that learns geometry-aware content features for stylization from a point cloud representation of the scene, resulting in high-quality stylized images that are consistent across views. Further, we introduce a novel training protocol to enable the learning using only 2D images. We demonstrate the superiority of our method via extensive qualitative and quantitative studies, and showcase key applications of our method in light of the growing demand for 3D content creation from 2D image assets.\footnote{Project page: \url{http://pages.cs.wisc.edu/~fmu/style3d}}

\end{abstract}

\section{Introduction}

Given an input content image and a reference style image, neural style transfer~\cite{gatys2016image,johnson2016perceptual,chen2017stylebank,huang2017arbitrary,li2017universal,ghiasi2017exploring,gu2018arbitrary,sheng2018avatar,park2019arbitrary,liu2021adaattn} creates a novel image that ``paints’’ the content with the style. Despite a high quality stylized image, the result is limited to the same viewpoint of the content image. What if we can render stylized images from different views? See Fig.\ \ref{fig:teaser} for examples. When displayed with parallax, this capacity will provide drastically more immersive visual experience for 2D images, and support the application of interactive browsing of 3D photos on mobile and AR/VR devices. 
In this paper, we address the new task of generating stylized images of novel views {\it from a single input image and an arbitrary reference style image}, as illustrated in Fig~\ref{fig:teaser}. We refer to this task as 3D photo stylization --- a marriage between style transfer and novel view synthesis.


3D photo stylization has several major technical barriers. As observed in~\cite{huang2021stylenvs}, directly combining existing methods of style transfer and novel view synthesis yields blurry or inconsistent stylized images, even with dense 3D geometry obtained from structure from motion and multi-view stereo. This challenge is further manifested with a single content image as the input, where a method must resort to monocular depth estimation with incomplete and noisy 3D geometry, leading to holes and artifacts when synthesizing stylized images of novel views. In addition, training deep models for this task requires a large-scale dataset of diverse scenes with dense geometry annotation that is currently lacking.

To bridge this gap, we draw inspiration from one-shot 3D photography~\cite{niklaus20193d,kopf2020one,shih20203d}, and adopt a point cloud based scene representation~\cite{niklaus20193d,wiles2020synsin,huang2021stylenvs}. Our key innovation is a deep model that learns 3D geometry-aware features on the point cloud {\it without using 2D image features from the content image} for rendering novel views with a consistent style. Our method accounts for the input noise from depth maps, and jointly models style transfer and view synthesis. Moreover, we propose a novel training scheme that enables learning our model using standard image datasets (\eg, MS-COCO~\cite{lin2014microsoft}), without the need of multi-view images or ground-truth depth maps. 

Our contributions are summarized into three folds. {\bf (1)} We present the first method to address the new task of 3D photo stylization --- synthesizing stylized novel views from a single content image with arbitrary styles. {\bf (2)} Unlike previous methods, our method learns geometry-aware features on a point cloud without using 2D content image features and from only 2D image datasets. {\bf (3)} Our method demonstrates superior qualitative and quantitative results, and supports several interesting applications.

\section{Related work}
\label{sec:related}

\noindent {\bf Neural Style Transfer}.
Neural style transfer has received considerable attention. Image style transfer~\cite{gatys2015neural,gatys2016image} renders the content of one image in the style of another. Video style transfer~\cite{ruder2018artistic} injects a style to a sequence of video frames to produce temporally consistent stylization, often by enforcing smoothness constraint on optical flow~\cite{huang2017real,chen2017coherent,ruder2018artistic,wang2020consistent} or in the feature space~\cite{deng2020arbitrary,liu2021adaattn}. Our method faces the same challenge as video style transfer; that the style must be consistent across views. However, our task of 3D photo stylization is more challenging, as it requires the synthesis of novel views and a consistent style among all views. 

Technically, early methods formulate style transfer as a slow iterative optimization process~\cite{gatys2015neural,gatys2016image}. Fast feed-forward models later perform stylization in a single forward pass, but can only accommodate one~\cite{johnson2016perceptual,ulyanov2016texture} or a few styles~\cite{dumoulin2016learned,chen2017stylebank}. Most relevant to our work are methods that allow for the transfer of {\it arbitrary} styles while retaining the efficiency of a feed-forward model~\cite{chen2016fast,huang2017arbitrary,li2017universal}. 
Our style transfer module builds on Liu~\etal~\cite{liu2021adaattn}, extending an attention-based method to support arbitrary 3D stylization. 

\noindent {\bf Novel View Synthesis from a Single Image}.
Novel view synthesis from a single image, also known as one-shot 3D photography, has seen recent progress thanks to deep learning. 
Existing approach can be broadly classified as end-to-end models~\cite{tulsiani2018layer,chen2019monocular,wiles2020synsin,tucker2020single,yu2021pixelnerf,rockwell2021pixelsynth,li2021mine,hu2021worldsheet} and modular systems~\cite{niklaus20193d,kopf2020one,shih20203d,jampani2021slide}. End-to-end methods 
often fail to recover accurate scene geometry and have difficulty generalizing beyond the scene categories present in training. Hence, our method builds on modular systems. 


Modular systems for one-shot 3D photography combine depth estimation~\cite{ranftl2020midas,wei2021leres,ranftl2021dpt} and inpainting models~\cite{liu2018image}, and have demonstrated strong results for in-the-wild images. Niklaus~\etal~\cite{niklaus20193d} maintains and rasterizes a point cloud representation of the scene to synthesize 3D Ken Burns effect. Later methods~\cite{kopf2020one,shih20203d} improve on synthesis quality via local content and depth inpainting on a layered depth image (LDI) of the scene. Jampani~\etal~\cite{jampani2021slide} further introduces soft scene layering to better preserve appearance details. Our work is closely related to Shih~\etal\cite{shih20203d}. We extend their LDI inpainting method for point cloud, and leverage their system to generate ``pseudo'' views during training. Our method also uses the differentiable rasterizer from~\cite{niklaus20193d}. 


\noindent {\bf 3D Stylization}. There has been a growing interest in the stylization of 3D content for creative shape editing~\cite{cao2020psnet,yin20213DStyleNet}, visual effect simulation~\cite{guo2021volumetric}, stereoscopic image editing~\cite{chen2018stereoscopic,gong2018neural} and novel view synthesis~\cite{huang2021stylenvs,chiang2021style3d}. Our method falls in this category and is most relevant to stylized novel view synthesis~\cite{huang2021stylenvs,chiang2021style3d}. The key difference is that our method generates stylized novel views from a single image, while previous methods need hundreds of calibrated views as input. Another difference is that our model learns 3D geometry aware features on a point cloud. In contrast, Huang~\etal~\cite{huang2021stylenvs} back-projects 2D image features to 3D space without accounting for scene geometry. While their point aggregation module enables {\it post hoc} processing of image-derived features, the point features remain 2D, leading to visual artifacts and inadequate stylization in renderings. Our work is also related to point cloud stylization \eg, PSNet~\cite{cao2020psnet} and 3DStyleNet~\cite{yin20213DStyleNet}. Both our method and~\cite{cao2020psnet,yin20213DStyleNet} use point cloud as the representation. The difference is that point cloud is an enabling device for stylization and view synthesis in our method, and not as the end product as in~\cite{cao2020psnet,yin20213DStyleNet}. 



\begin{figure*}
    \centering
    \resizebox{0.9\textwidth}{!}{
    \includegraphics{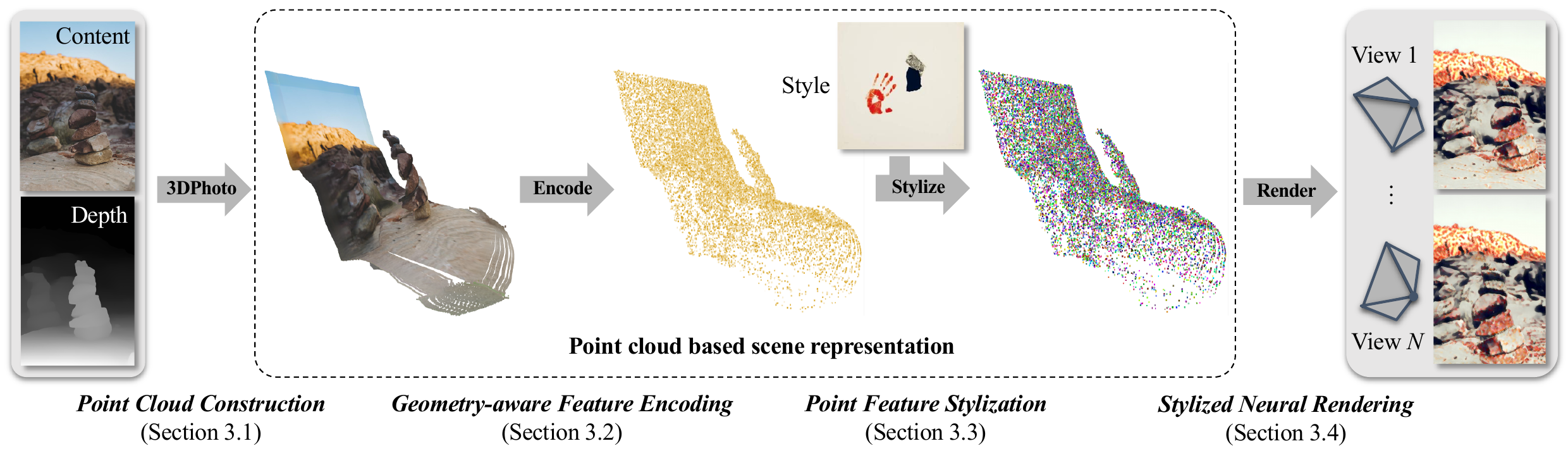}}\vspace{-1em}
    \caption{{\bf Method overview. } Central to our method is a point cloud based scene representation that enables geometry-aware feature learning, attention-based feature stylization and consistent stylized renderings across views. Specifically, we first construct an RGB point cloud from the content image and its estimated depth map. Content features are then extracted directly from the point cloud and stylized given an image of the reference style. Finally, the stylized point features are rendered to novel views and decoded into stylized images.}
    \label{fig:workflow}
\vspace{-0.2in}
\end{figure*}

\noindent {\bf Deep Models for Point Cloud Processing}. 
Many deep models have been developed for point cloud processing.
Among the popular architectures are models of set based \cite{qi2017pointnet,qi2017pointnet++},
graph convolution based \cite{wang2019dgcnn,li2021deepgcns_pami} and point convolution based \cite{hua2018pointwise,thomas2019KPConv}. 
Our model extends a graph based model~\cite{wang2019dgcnn} to handle dense point clouds 
(one million points) for high quality stylization. 

\section{3D Photo Stylization}
\label{sec:inference}

Given {\it a single input content image} and {\it an arbitrary style image}, the goal of 3D photo stylization is to generate stylized novel views of the content image.
The key of our method is the learning of 3D geometry aware content features directly from a point cloud representation of the scene for high-quality stylization that is consistent across views. 
In this section, we describe our workflow at {\it inference} time.



\noindent {\bf Method Overview}.
Fig.\ \ref{fig:workflow} presents an overview of our method. Our method starts by back-projecting the input content image into an RGB point cloud using its estimated depth map. The point cloud is further ``inpainted'' to cover dissoccluded parts of the scene and then ``normalized'' (Section~\ref{subsec:construct}). An efficient graph convolutional network is designed to process the point cloud and extract 3D geometry aware features on the point cloud, leading to point-wise features tailored for 3D stylization (Section~\ref{subsec:encode}). A style transfer module is subsequently adapted to modulate those point-wise features using the input style image (Section~\ref{subsec:stylize}). Finally, a differentiable rasterizer projects the featurized points to novel views for the synthesis of stylized images that are consistent across views (Section~\ref{subsec:render}).


\subsection{Point Cloud Construction}
\label{subsec:construct}

Our method starts by lifting the content image into an RGB point cloud, and further normalizes the point cloud to account for scale ambiguity and uneven point density.


\noindent {\bf Depth Estimation and Synthesis of Hidden Geometry}.
Our method first estimates a dense depth map using an off-the-shelf deep model for monocular depth estimation (LeReS~\cite{wei2021leres}). A key challenge for single-image novel view synthesis is the occlusion in the scene. A dense depth map might expose many ``holes'' when projected to a different view. Inpainting the occluded geometry is thus critical for view synthesis. To this end, we further employs the method of Shih~\etal~\cite{shih20203d} for the synthesis of occluded geometry on a layered depth image (LDI). Thanks to the duality between point cloud and LDI, we map the LDI pixels to an RGB point cloud via perspective back-projection. 

\begin{figure}
    \centering
    \includegraphics[width=1.0\linewidth]{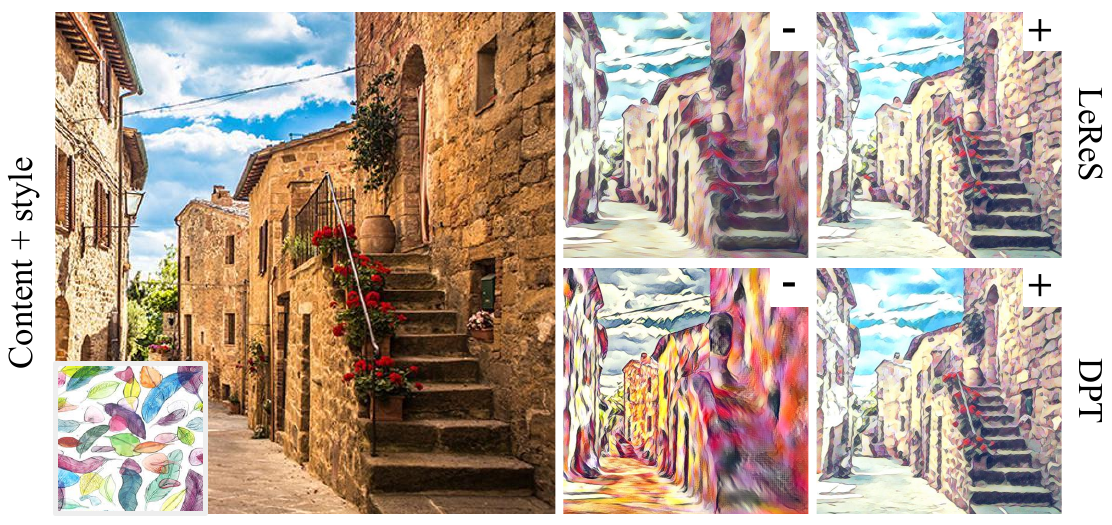}\vspace{-0.5em}
    \caption{{\bf Effect of point cloud normalization. } Model without normalization (-) performs poorly due to scale ambiguity in depth estimation and non-uniformity in point distribution. In contrast, model with normalization (+) captures fine appearance detail and produces strong stylization irrespective of depth estimator in use.}
    \label{fig:normalization}
\vspace{-1.5em}
\end{figure}

\noindent {\bf Point Cloud Normalization}. In light of scale ambiguity and uneven point density characteristic of image-derived point clouds, we transform them into Normalized Device Coordinate (NDC)~\cite{marschner2021fundamentals} before further processing. 
The resulting points fall within the $[-1, 1]$ cube with density adjusted accordingly to account for perspectivity. As shown in Fig~\ref{fig:normalization}, this simple procedure is crucial for our method to generalize across scene categories, and allows us to switch to different depth estimators without re-training our model.

\subsection{Encoding Features on Point Cloud}
\label{subsec:encode}

Our next step is to learn features amenable to stylization.
While virtually all existing style transfer algorithms make use of ImageNet pre-trained VGG features, we found that associating 3D points with back-projected VGG features (such as in Huang~\etal~\cite{huang2021stylenvs}) is sub-optimal for stylized novel view synthesis, leading to geometric distortion and structural artifacts as shown in our ablation. We argue that features from a network pre-trained on 2D images are incompetent to describe the intricacy of 3D geometry. This leads us to design an efficient graph convolutional network (GCN) that learns geometry aware features directly from an RGB point cloud, as opposed to using 2D image features. 

\begin{figure*}
    \centering \vspace{-0.4em}
    \resizebox{0.9\textwidth}{!}{
    \includegraphics{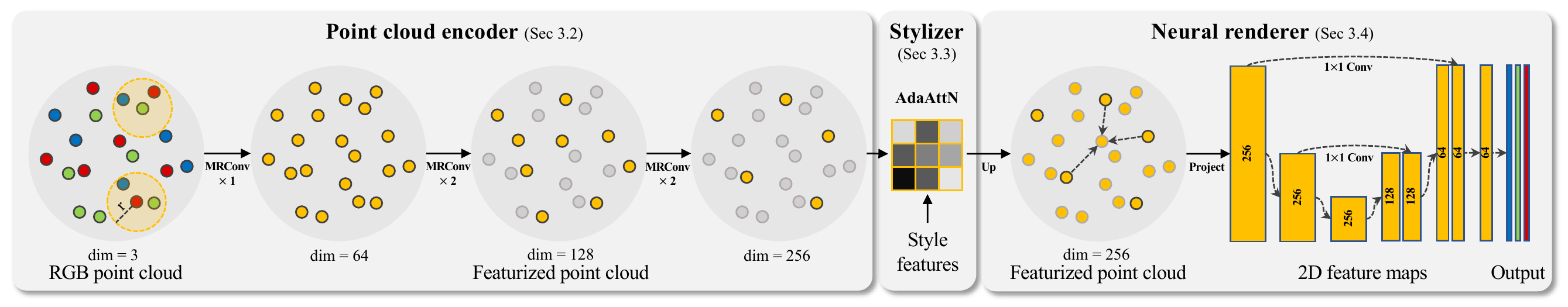}}\vspace{-1em}
    \caption{{\bf Components of our deep model. } Our model includes three modules --- a point cloud encoder, a stylizer and a neural renderer. The encoder applies MRConvs~\cite{li2021deepgcns_pami} along with farthest point sampling to embed and sub-sample the input RGB point cloud. The stylizer computes attention between the embedded content and style features, and uses attention-weighted affine transformation to modulate the content features for stylization. The neural render consists of a rasterizer that anti-aliases the modulated point features and projects them to novel views, and a U-Net~\cite{ronneberger2015unet} that refines the resulting 2D feature maps and decodes them into stylized images.}
    \label{fig:technical}
\vspace{-1.5em}
\end{figure*}


\noindent \textbf{Efficient GCN}. One common drawback for GCN architectures lies in their scalability.
Existing GCNs are designed for points clouds with a few thousand points~\cite{li2021deepgcns_pami}, whereas an image at 1K resolution results in {\it one million} points after inpainting. 
To bridge this gap, we propose a highly efficient GCN encoder by drawing strength from multiple point-based network architectures.

Our GCN encoder adopts the max-relative convolution~\cite{li2021deepgcns_pami} for its computational and memory efficiency. To further improve the efficiency, we replace the expensive dynamic k-NN graphs with radius-based ball queries~\cite{qi2017pointnet++} for point aggregation. Moreover, we follow the hierarchical design of VGG network by repeatedly sub-sampling the point cloud via farthest point sampling, as opposed to maintaining the full set of points throughout the model~\cite{li2021deepgcns_pami}. 
We illustrate our encoder design in Fig.\ \ref{fig:technical}. The output of our encoder is a sub-sampled, featurized point cloud. 

\subsection{Stylizing the Point Cloud}
\label{subsec:stylize}
Going further, our model injects style into the content features. The technical barrier here is the misalignment of content and style features, as the former are defined on a 3D point cloud while the latter (from a pre-trained VGG network) lie in a 2D plane. To address this discrepancy, we make use of learned feature mappings and Adaptive Attention Normalization (AdaAttN)~\cite{liu2021adaattn} to match and combine the content and style features. Let $F_{c}$ be the point-wise content features and $F_{s}$ the style features on a 2D grid. Our style transfer operation is given by 
\begin{equation}
\small
    F_{cs} = \psi(\mbox{AdaAttN}(\phi(F_{c}), F_{s})),
\end{equation}
where $\phi$ and $\psi$, implemented as point-wise multi-layer perceptrons (MLPs), are learned mappings between the content and style feature spaces, and $\mbox{AdaAttN}$ is the attention-weighted adaptive instance normalization from~\cite{liu2021adaattn}. $\mbox{AdaAttN}$ computes attention between every content feature (a point) and each style feature (a pixel), and uses the attention map to modulate the affine parameters within the instance normalization applied on content features. As a result, $F_{cs}$ incorporates both content and style, and will be further used to render stylized images.

\subsection{Stylized Neural Rendering}
\label{subsec:render}

Our final step is to render stylized point features $F_{cs}$ into stylized images with specified viewpoints.
As illustrated in Fig~\ref{fig:technical}, this is accomplished by (1) projecting point features to an image plane given camera pose and intrinsics; and (2) decoding the projected features into an image using a 2D convolutional network.

\noindent {\bf Feature Rasterization}.
Our rasterizer follows  Niklaus~\etal~\cite{niklaus20193d}, and projects the point cloud features $F_{cs}$ into a single-view 2D feature map $F_{2d}$. There is one important difference: we up-sample $F_{cs}$ using inverse distance weighted interpolation~\cite{qi2017pointnet++} {\it before rasterization}. This is reminiscent of super-sampling --- a classical anti-aliasing technique in graphics. In doing so, we grant more flexibility for decoding the projected features into stylized images. 

\noindent {\bf Image Decoding}.
Our decoder further maps the 2D feature map $F_{2d}$ to a stylized RGB image at input resolution. The decoder is realized using a 2D convolutional network, following the architecture of U-Net~\cite{ronneberger2015unet}, with transposed convolutions at the entry of each stage for up-sampling.

\section{Learning from 2D Images}
\label{sec:train}

We now present our training scheme. Our model is trained using 2D images following a two-stage approach. 

\begin{figure*}
    \centering \vspace{-1em}
    \includegraphics[width=0.9\linewidth]{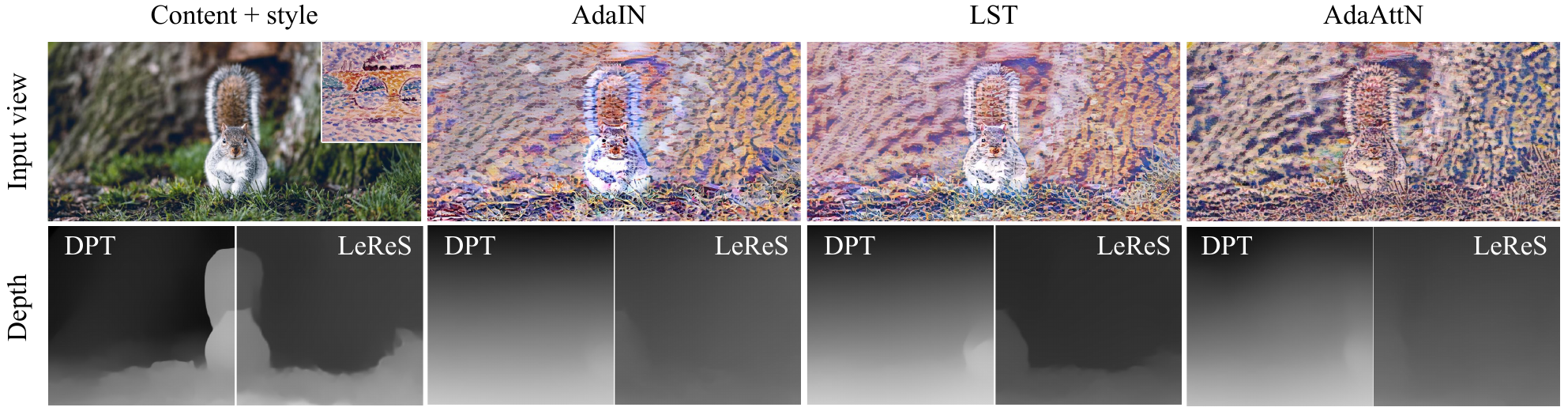}\vspace{-0.8em}
    \caption{{\bf Depth estimation fails on stylized images. } One alternative to 3D photo stylization is to combine stylized content image and its depth estimate. Unfortunately, strong depth estimators such as DPT~\cite{ranftl2021dpt} and LeReS~\cite{wei2021leres} fail on image style transfer output from AdaIN~\cite{huang2017real}, LST~\cite{li2019learning} and AdaAttN~\cite{liu2021adaattn} because stylized images do not follow natural image statistics.}\vspace{-1.2em}
    \label{fig:baseline1_1}
\end{figure*}

\begin{figure*}
    \centering
    \includegraphics[width=0.9\linewidth]{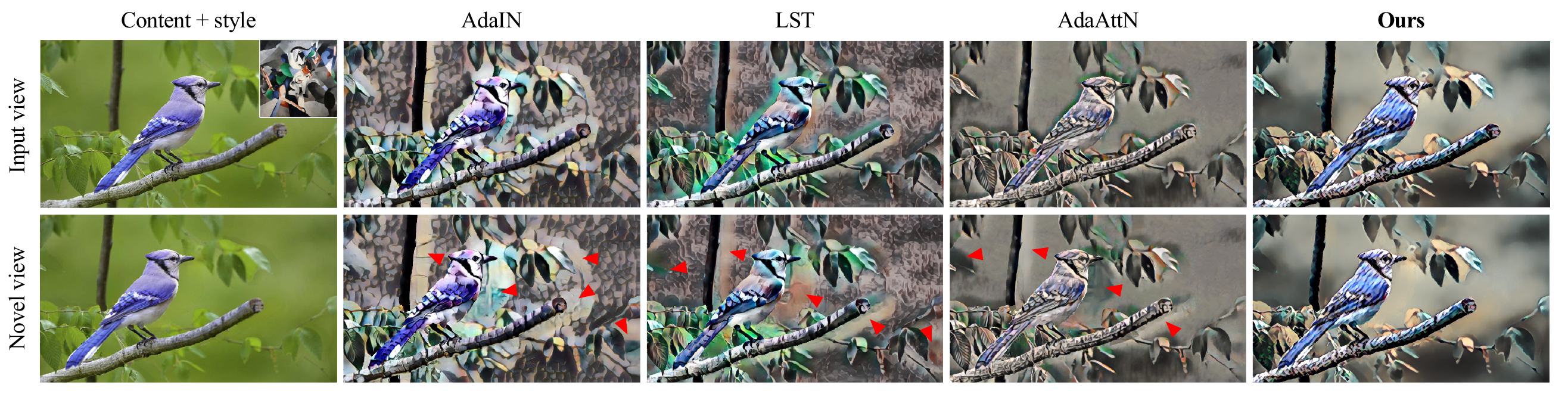}\vspace{-1em}
    \caption{{\bf 3D photo of a stylized content image manifests ubiquitous visual artifacts. } Another alternative to stylizing 3D photos is to combine stylized content image with depth estimate from the {\it original} content image. While depth estimation is unaffected, the style effect bleeds through depth discontinuities. 3D photo inpainting thus fails, with ubiquitous visual artifacts (\textcolor{red}{{\bf red}} arrows) in novel view renderings.}\vspace{-1.5em}
    \label{fig:baseline1_2}
\end{figure*}

\noindent \textbf{Generating Multi-view Images for Training}. Training our model requires images from multiple views of the same scene. Unfortunately, a large-scale multi-view image dataset with a diverse set of scenes is lacking. To bridge this gap, we propose to learn from the results of existing one-shot 3D photography methods. 
Concretely, we use 3DPhoto~\cite{shih20203d} to convert images from a standard dataset (MS-COCO) into high-quality 3D meshes, from which we synthesize {\it arbitrary} pseudo target views to train our model. In doing so, our model learns from a diverse collection of scenes present in MS-COCO. Learning from synthesized images leads to an inevitable bias residing in 3DPhoto results in trade of dataset diversity. Through our experiments, we show that our model generalizes well across a large set of in-the-wild images at inference time.


\subsection{Two-Stage Training}
The training of our model is divided into a {\it view synthesis} stage where the model learns 3D geometry aware features for novel view synthesis, and a {\it stylization} stage where the model is further trained for novel view stylization.

\noindent {\bf Enforcing Multi-view Consistency}.
A key technical contribution of our work is a multi-view consistency loss. Building a point cloud representation of the input content image allows us to impose additional constraint on {\it pixel values} of the rendered images.\footnote{While the sharing of a featurized point cloud entails multi-view consistency of rasterized {\it feature maps}, the features are subject to a learnable decoding process, through which inconsistency will be introduced.} The key idea is that a scene point $\mathbf{p}$ in the point cloud $\mathbf{P}$ should produce the same pixel color in the views to which it is visible. To this end, we define our consistency loss as
\begin{equation}
\small
    \mathcal{L}_{cns} = \sum_{\mathbf{p} \in \mathbf{P}}\sum_{i,j \in \mathbf{V}} \mathcal{V}(p; i, j) \cdot \|\mathbf{I}_{i}(\pi_{i}(\mathbf{p})) - \mathbf{I}_{j}(\pi_{j}(\mathbf{p}))\|_{1},
\end{equation}
where $\mathbf{V}$ is the set of sampled views, $\mathbf{I}_i$ the rendered image from view $i$, $\pi_{i}(\cdot)$ the projection to view $i$, and $\mathcal{V}(p; \cdot, \cdot)$ a visibility function which evaluates to 1 if $p$ is visible to both views and 0 otherwise. Computing the loss incurs minimal overhead since the evaluation of $\pi$ and $\mathcal{V}$ is part of rasterization. As evidenced by our ablation study, our proposed loss significantly improves consistency of stylized renderings.

\noindent {\bf View Synthesis Stage}.
We first train our model for view synthesis, a surrogate task that drives the learning of geometry aware content features. Given an input image, we randomly sample novel views of the scene and ask the model to reconstruct them. To train our model, we make use of an L1 loss $\mathcal{L}_{rgb}$ defined on pixel values, a VGG perceptual loss $\mathcal{L}_{feat}$ defined on network features, and our multi-view consistency loss $\mathcal{L}_{cns}$. The overall loss function is 
\begin{equation}
\small
    \mathcal{L}_{view} = \mathcal{L}_{rgb} + \mathcal{L}_{feat} + \mathcal{L}_{cns},
\end{equation}

\noindent {\bf Stylization Stage}.
Our model learns to stylize novel views in the second stage. We freeze the encoder for content feature extraction, train the stylizer, and fine-tune the neural renderer. This is done by randomly sampling novel views of the scene and style images from WikiArt~\cite{nichol2016wikiart}, and training our model using 
\begin{equation}
\small
    \mathcal{L}_{style} = \mathcal{L}_{adaattn} + \mathcal{L}_{cns},
\end{equation}
where $\mathcal{L}_{adaattn}$ is the same AdaAttN loss from \cite{liu2021adaattn} and $\mathcal{L}_{cns}$ is again our multi-view consistency loss. 

\noindent {\bf Training Details}.
For view synthesis, we train for 20K iterations (2 epochs) on MS-COCO with a batch size of 8 using Adam~\cite{kingma2015adam} and set the learning rate to 1e-4. We apply the same training schedule for stylization. 

\begin{figure*}
    \centering \vspace{-1em}
    \resizebox{0.95\textwidth}{!}{
    \includegraphics{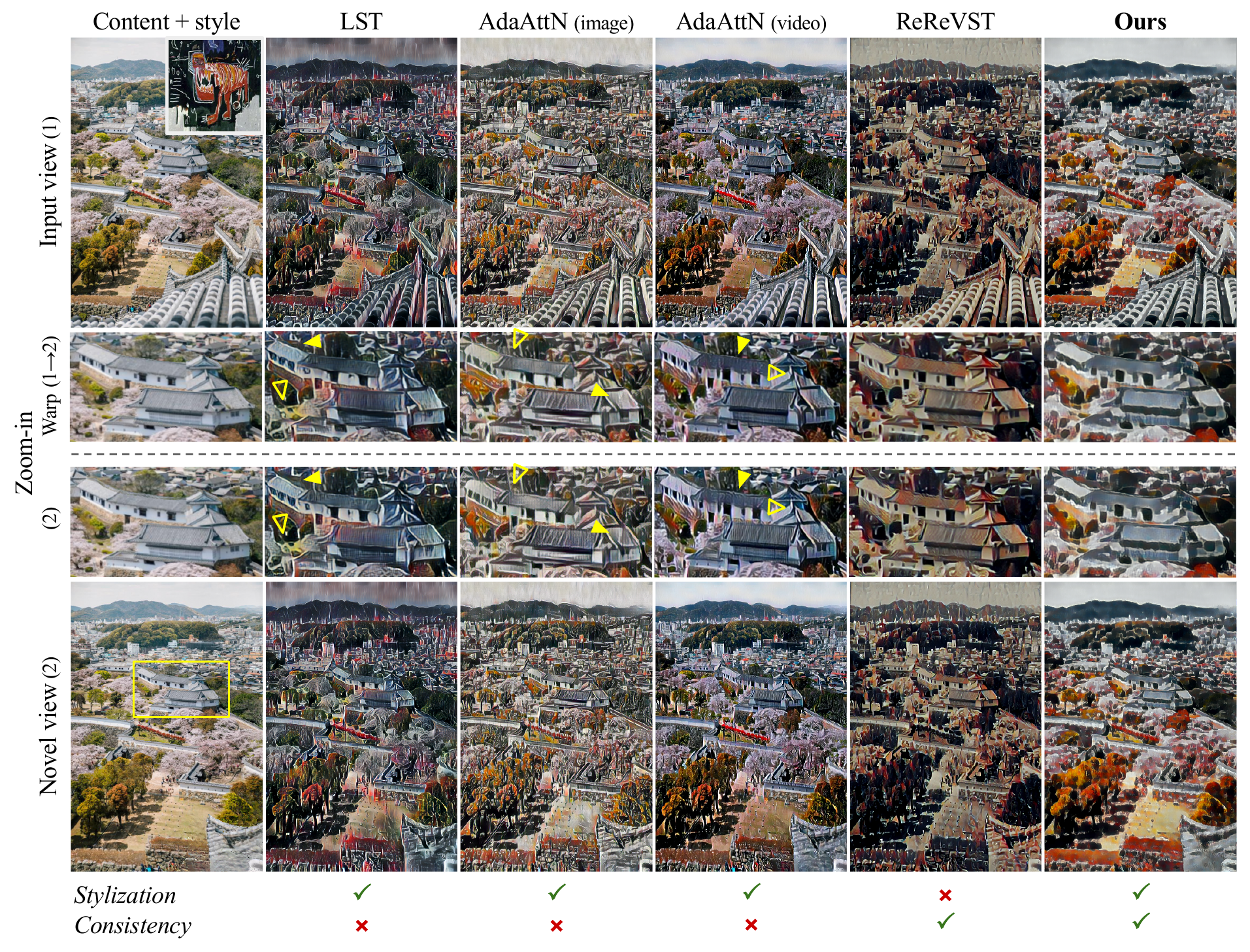}}\vspace{-1.2em}
    \caption{{\bf Stylizing rendered images from a 3D photo introduces inconsistency in stylization. } A third baseline is to na\"ively build a 3D photo from the raw content image, then stylize its renderings either one view at a time (\eg, using LST~\cite{li2019learning} or AdaAttN~\cite{liu2021adaattn}) or collectively as a video (\eg, using ReReVST~\cite{wang2020consistent} or the video variant of AdaAttN). Despite stronger results than the other two baselines, the stylization is agnostic to the scene geometry shared by all views and thus produces inconsistent results (\textcolor{yellow}{{\bf yellow}} arrows).}\vspace{-1.5em}
    \label{fig:baseline2}
\end{figure*}

\section{Experiments}
\label{sec:result}

We now present the main results of our paper and leave additional results to the supplementary material. 


\subsection{Qualitative results}
\label{subsec:qualitative}

By permuting the steps of (1) depth estimation, (2) inpainting, (3) rendering and (4) style transfer, one could imagine two alternative workflows that combine existing models for 3D photo stylization. To compare them with our method, we instantiate these baselines by combining six different style transfer methods (AdaIN~\cite{huang2017arbitrary}, LST~\cite{li2019learning} and AdaAttN~\cite{liu2021adaattn} for image style transfer, and ReReVST~\cite{wang2020consistent}, MCC~\cite{deng2020arbitrary} and the video variant of AdaAttN for video style transfer) with DPT~\cite{ranftl2021dpt} for depth estimation and 3DPhoto~\cite{shih20203d} for inpainting and rendering. Results are created using images from Unsplash~\cite{unsplash2020}, a free-licensed, professional-grade dataset of in-the-wild images.

(1) {\it Style $\rightarrow$ Depth $\rightarrow$ Inpainting $\rightarrow$ Rendering}: 
While geometric consistency is granted, depth estimation fails catastrophically on stylized images (Figure~\ref{fig:baseline1_1}). One may alternatively back-project a stylized image using depth estimation from the raw input. Despite better geometry, inpainting remains error-prone due to color bleed-through and shift in color distribution caused by stylization (Figure~\ref{fig:baseline1_2}).

(2) {\it Depth $\rightarrow$ Inpainting $\rightarrow$ Rendering $\rightarrow$ Style}:
This baseline often produces inconsistent stylization across views (Figure~\ref{fig:baseline2}), as each view's style is independent and agnostic to the underlying scene geometry.

In contrast, our method manages to generate high-quality stylized renderings free of visual artifacts and inconsistency. The second baseline produces gentle inconsistency under small viewpoint change typical to 3D photo browsing. This is more benign than the visual artifacts produced by the first baseline. We further compare our method with the second baseline via quantitative experiments and a user study.

\subsection{Quantitative results}
\label{subsec:quantitative}
Given that evaluation of style quality is a very subjective matter, we defer it to the user study and focus on the evaluation of consistency in our quantitative experiments.

\noindent {\bf Evaluation Protocol and Metrics}.
We run our method and the baseline on ten diverse content images from the web and 40 styles sampled from the compilation of Gao~\etal~\cite{gao2020fast}. The baseline, as discussed before, runs 3DPhoto to synthesize {\it plain} novel-view images, then stylizes them using one of the six style transfer algorithms. Ultimately, this results in 400 stylized 3D photos from each of the seven candidate methods. To quantify inconsistency between a pair of stylized views, we warp one view to the other according to the point cloud based scene geometry, and compute RMSE and the masked LPIPS metric as defined in Huang~\etal~\cite{huang2021stylenvs}. We average the result over 400 pairs of views for each stylized 3D photo and report the mean over all available photos.

\noindent {\bf Results}.
Our results are summarized in Table~\ref{tab:cns}. Our method outperforms all six instantiations of the baseline by a significant margin in terms of both RMSE and LPIPS. Not surprisingly, video style transfer methods produce more consistent results than image style transfer methods owing to their extra smoothness constraint. The fact that our method performs even better without such a constraint shows the effectiveness of maintaining a central featurized point cloud for 3D photo stylization.

\begin{table}[t]
\centering
\resizebox{0.8\columnwidth}{!}{
\begin{tabular}{ll||cc} 
\hline
\multicolumn{2}{c||}{Method}                & \Gape[10pt][10pt]{RMSE}           & LPIPS           \\
\hline
\multirow{6}{*}{3DPhoto~\cite{shih20203d} $\rightarrow$} & AdaIN~\cite{huang2017arbitrary}           & 0.222          & 0.304           \\
                          & LST~\cite{li2019learning}             & 0.195          & 0.287           \\
                          & AdaAttN (image)~\cite{liu2021adaattn} & 0.187          & 0.329           \\ 
\cline{2-4}
                          & ReReVST~\cite{wang2020consistent}         & 0.115          & 0.213           \\
                          & MCC~\cite{deng2020arbitrary}             & 0.092          & 0.200           \\
                          & AdaAttN (video)~\cite{liu2021adaattn} & 0.135          & 0.209           \\ 
\hline
\multicolumn{2}{l||}{\textbf{Ours}}                  & \textbf{0.086} & \textbf{0.133}  \\
\hline
\end{tabular}}\vspace{-0.5em}
\caption{{\bf Results on consistency. } We compare our model against baselines that sequentially combine 3DPhoto and image/video style transfer on consistency using RMSE ($\downarrow$) and LPIPS ($\downarrow$).}
\vspace{-1em}
\label{tab:cns}
\end{table}

\begin{figure}[t!]
    \centering
    \resizebox{0.9\columnwidth}{!}{
    \includegraphics{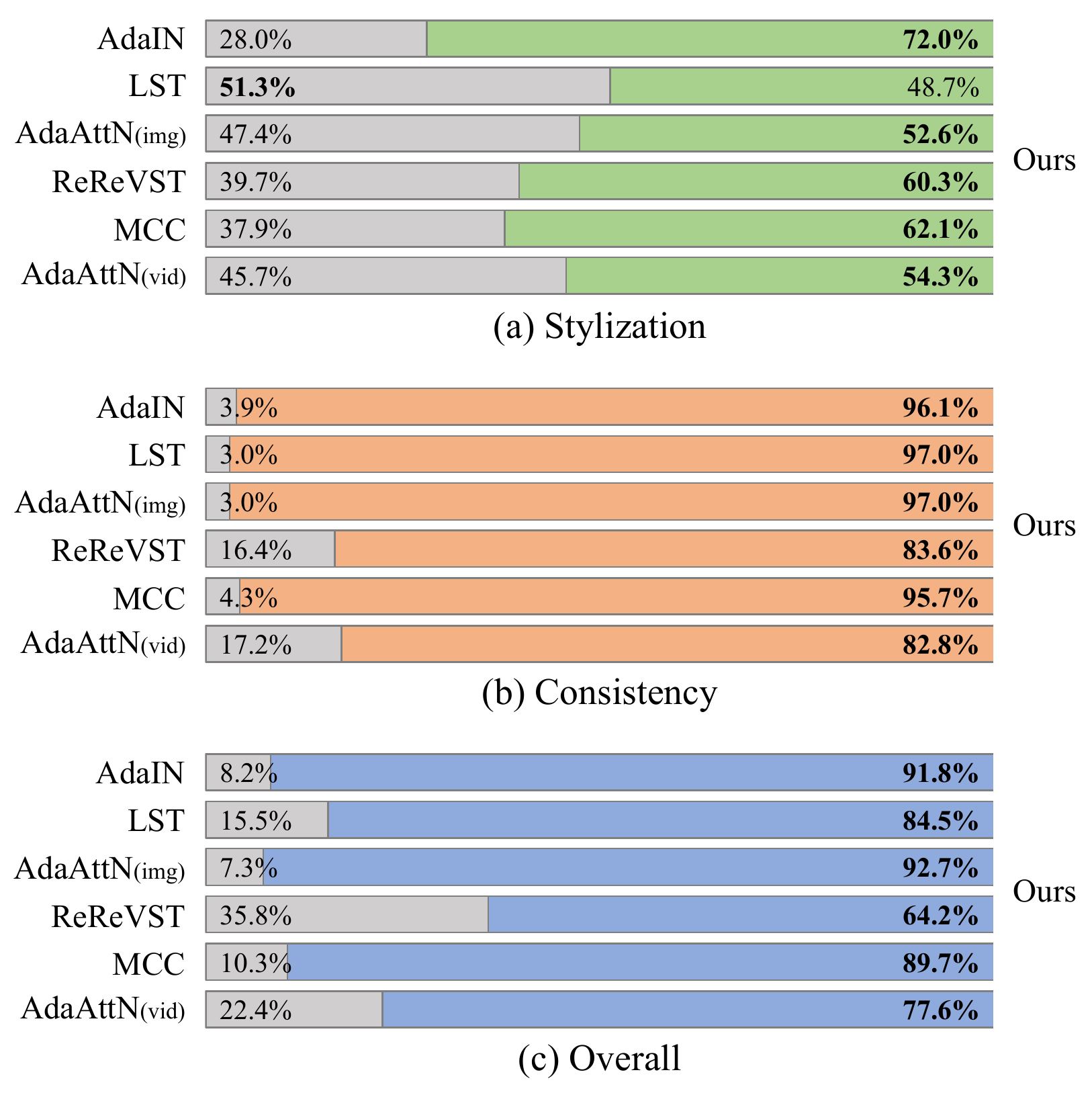}}\vspace{-1em}
    \caption{{\bf User study. } We conduct a user study to compare our method against baselines that sequentially combine 3DPhoto and image/video style transfer. Methods are evaluated on (a) style quality, (b) multi-view consistency and (c) overall synthesis quality. Results show percentage of users voting for an algorithm.}
    \label{fig:user}
\vspace{-1.5em}
\end{figure}

\subsection{User study}
\label{subsec:user}
Going further, we conduct a user study to better understand the perceptual quality of stylized images produced by our method and the baselines. Our study includes three sections for the assessment of style quality, multi-view consistency and overall synthesis quality. Our analysis is based on 5,400 votes from 30 participants. We elaborate on our study design in the supplementary material.


\noindent {\bf Results}.
We visualize the results in Figure~\ref{fig:user}. For style quality, our method is consistently rated better than the alternatives, with the only exception being LST, which our method is on par with. Not coincidentally, our method excels at multi-view consistency, harvesting an overwhelming 95 percent of the votes in four of the six tests. Finally, our method remains the most preferred for overall synthesis quality, beating all alternatives by a large gap. Putting things together, our results provide solid validation on the strength of our approach in producing high-quality stylization that is consistent across views.

\begin{figure}
    \centering \vspace{-0.5em}
    \includegraphics[width=1.0\linewidth]{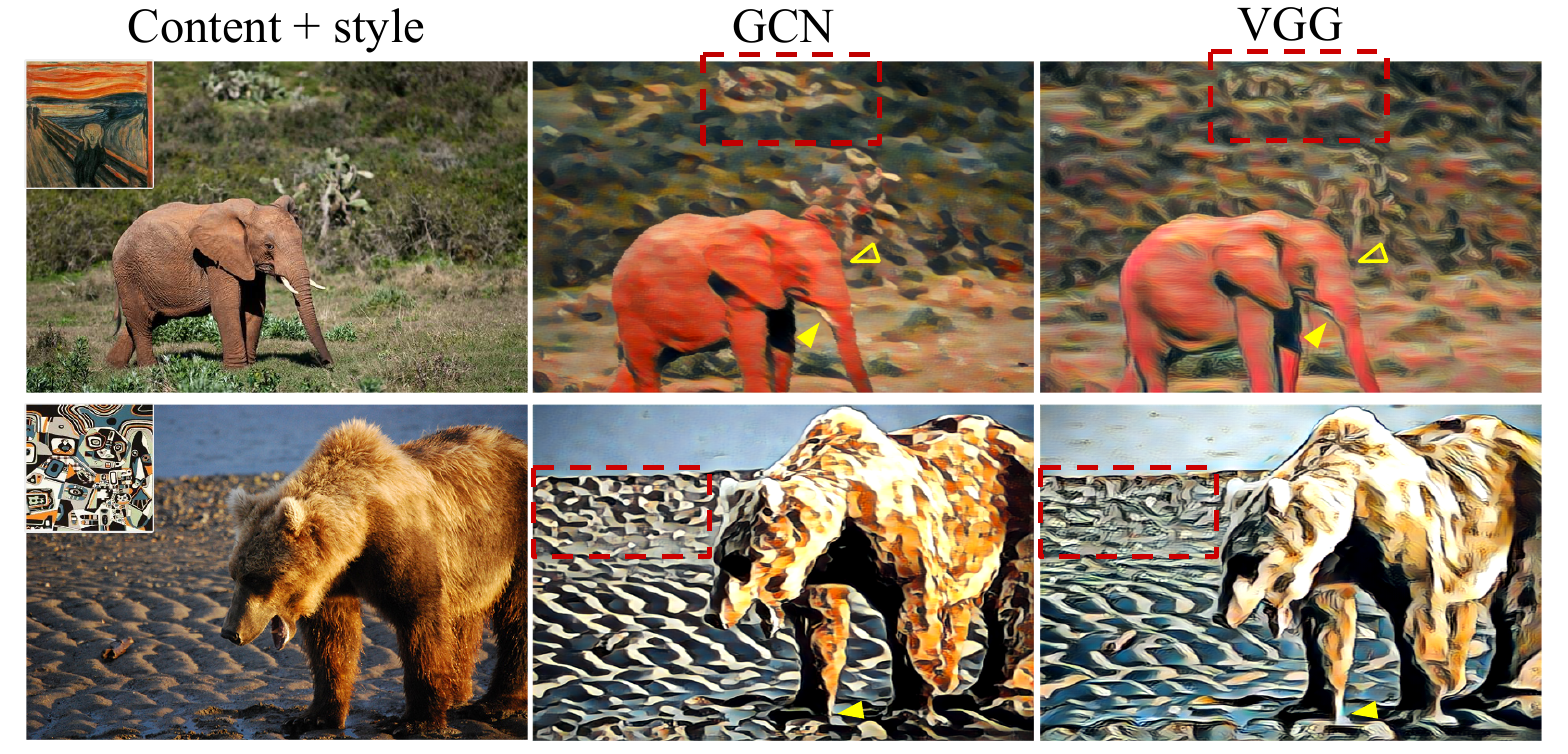}\vspace{-1em}
    \caption{{\bf Effect of geometry-aware feature learning. } 3D photo stylization with back-projected 2D VGG features suffers from geometric distortion (\textcolor{yellow}{{\bf yellow}} arrows) and visual artifacts (\textcolor{red}{{\bf red}} boxes). In contrast, our geometry-aware learning scheme better maintains content structure and produces more pleasant texture.}\vspace{-0.5em}
    \label{fig:geometry}
\end{figure}

\begin{table}
\centering
\resizebox{0.55\linewidth}{!}{
\begin{tabular}{cc||cc} 
\hline
\multicolumn{2}{c||}{Training stage}                             & \multicolumn{1}{l}{\multirow{2}{*}{RMSE}} & \multicolumn{1}{l}{\multirow{2}{*}{LPIPS}}  \\
\textit{ViewSyn}                              & \textit{Stylize} & \multicolumn{1}{l}{}                      & \multicolumn{1}{l}{}                        \\ 
\hline
\begin{tabular}[c]{@{}c@{}}$-$\\\end{tabular} & $-$              & 0.113                                     & 0.199                                       \\
\begin{tabular}[c]{@{}c@{}}$+$\\\end{tabular} & $-$              & 0.109                                     & 0.190                                       \\
\begin{tabular}[c]{@{}c@{}}$-$\\\end{tabular} & $+$              & \textbf{0.081}                            & 0.132                                       \\
\begin{tabular}[c]{@{}c@{}}$+$\\\end{tabular} & $+$              & 0.086                                     & \textbf{0.128}                              \\
\hline
\end{tabular}}\vspace{-0.5em}
\caption{{\bf Effect of consistency loss. } We compare models trained with (+) or without (-) the loss using RMSE ($\downarrow$) and LPIPS ($\downarrow$).}
\vspace{-1.5em}
\label{tab:abcns}
\end{table}

\subsection{Ablation studies}
\label{subsec:ablation}


\smallskip
\noindent {\bf Effect of Geometry-aware Feature Learning}.
We study the strength of geometry-aware feature learning. Specifically, we construct a variant of our model with the only difference that content features are not learned on the point cloud, but rather come from a pre-trained VGG network as in 2D style transfer methods. In particular, we sidestep our proposed GCN encoding scheme by projecting an RGB point cloud to eight extreme views defined by a bounding volume, running the VGG encoder for feature extraction, and back-projecting the 2D features to a point cloud from which stylization and rendering proceed as before. As shown in Fig~\ref{fig:geometry}, this VGG-based variant produces geometric distortion and visual artifacts in stylized images, as opposed to our model using geometry-aware feature learning.

\begin{figure*}[t]
    \centering \vspace{-1em}
    \includegraphics[width=0.95\linewidth]{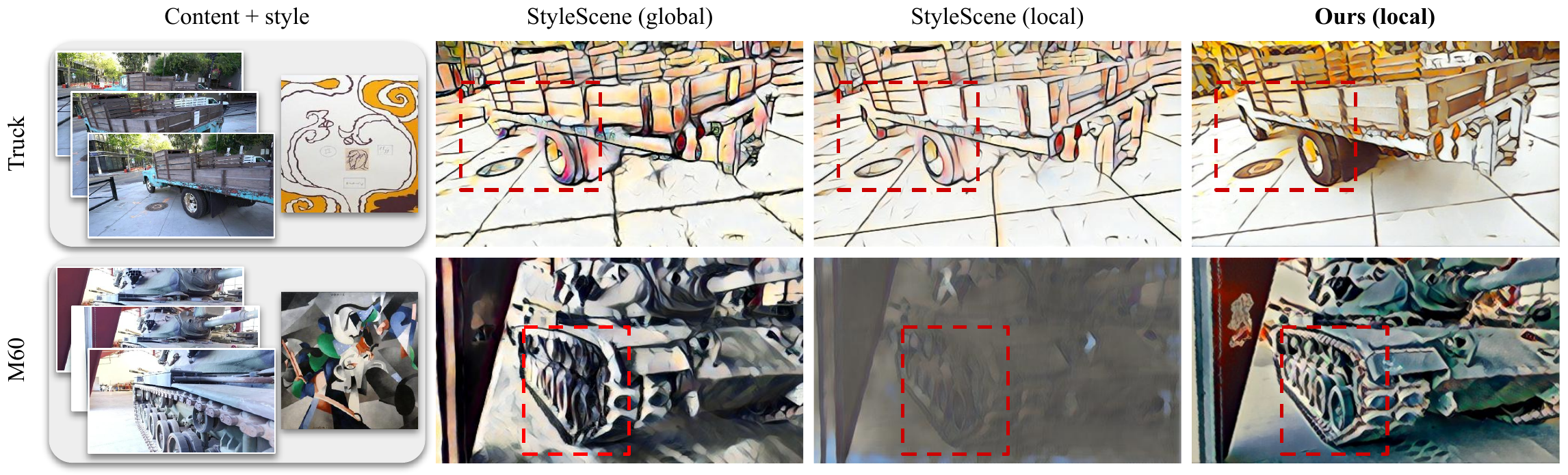}\vspace{-1em}
    \caption{{\bf Extension to multi-view input. } Compared with StyleScene~\cite{huang2021stylenvs}, our method more closely resembles the reference style, better preserves the content geometry (\textcolor{red}{{\bf red}} boxes), and is more robust to change in viewpoint distribution (second row).}
    \label{fig:extension}
\vspace{-0.5em}
\end{figure*}

\begin{table*}[t]
\centering
\resizebox{1.0\textwidth}{!}{
\begin{tabular}{c||cc|cc|cc|cc||cc|cc|cc|cc} 
\hline
\multirow{3}{*}{Method}                   & \multicolumn{8}{c||}{Short-range consistency}                                                                                                             & \multicolumn{8}{c}{Long-range consistency}                                                                                                               \\ 
\cline{2-17}
                                          & \multicolumn{2}{c|}{\textit{Truck}} & \multicolumn{2}{c|}{\textit{Playground}} & \multicolumn{2}{c|}{\textit{Train}} & \multicolumn{2}{c||}{\textit{M60}} & \multicolumn{2}{c|}{\textit{Truck}} & \multicolumn{2}{c|}{\textit{Playground}} & \multicolumn{2}{c|}{\textit{Train}} & \multicolumn{2}{c}{\textit{M60}}  \\
                                          & RMSE           & LPIPS              & RMSE           & LPIPS                   & RMSE           & LPIPS              & RMSE           & LPIPS             & RMSE           & LPIPS              & RMSE           & LPIPS                   & RMSE           & LPIPS              & RMSE           & LPIPS            \\ 
\hline
\multicolumn{1}{l||}{StyleScene (global)} & 0.124          & 0.143              & 0.108          & 0.142                   & 0.121          & 0.157              & 0.120          & 0.143             & 0.163          & 0.188              & 0.146          & 0.189                   & 0.159          & 0.213              & 0.160          & 0.192            \\
\multicolumn{1}{l||}{StyleScene (local)}  & 0.119          & 0.168              & 0.127          & 0.169                   & 0.161          & 0.169              & N/A            & N/A               & 0.152          & 0.203              & 0.166          & 0.205                   & 0.204          & 0.220              & N/A            & N/A              \\
\hline
\multicolumn{1}{l||}{\textbf{Ours (local)}}        & \textbf{0.099} & \textbf{0.107}     & \textbf{0.093} & \textbf{0.111}          & \textbf{0.104} & \textbf{0.112}     & \textbf{0.117} & \textbf{0.112}    & \textbf{0.113} & \textbf{0.128}     & \textbf{0.110} & \textbf{0.127}          & \textbf{0.120} & \textbf{0.145}     & \textbf{0.136} & \textbf{0.136}   \\
\hline
\end{tabular}}\vspace{-0.5em}
\caption{{\bf Consistency in the multi-view scenario. } On the Tanks and Temples dataset~\cite{knapitsch2017tanks}, we compare our method with StyleScene on short- and long-range consistency as defined in~\cite{huang2021stylenvs} using RMSE ($\downarrow$) and LPIPS ($\downarrow$).}\vspace{-1.5em}
\label{tab:extension}
\end{table*}

\noindent {\bf Effect of Consistency Loss}.
We evaluate the contribution of our consistency loss in Table~\ref{tab:abcns}. Despite a shared point cloud, model trained without the consistency loss produces less consistent renderings measured in RMSE and LPIPS. We attribute this to the learnable feature decoding step, which is too flexible to preserve consistency in output images in the absence of a constraint. In this respect, our consistency loss, especially when applied in the stylization stage of training, acts as a strong regularizer on the decoder.


\subsection{Extension to Multi-view Inputs}
\label{subsec:extension}
Our method can be easily extended for stylized novel view synthesis given multi-view inputs. We compare our extension with StyleScene~\cite{huang2021stylenvs}, which similarly operates on point cloud but requires multiple input views. We perform experiments on the Tanks and Temples dataset~\cite{knapitsch2017tanks} under two protocols. The {\it global} protocol uses all available views (up to 300) as in~\cite{huang2021stylenvs} for point cloud reconstruction, whereas the more challenging {\it local} protocol uses a sparse set of 6-8 views on the camera trajectory for novel view synthesis. In Fig~\ref{fig:extension} and Table~\ref{tab:extension}, we show that our method is better in terms of style quality, short- and long-range consistency, and robustness to the distribution of input views.

\subsection{Applications}
\label{sec:application}

\begin{figure}[thp]
    \centering \vspace{-0.5em}
    \includegraphics[width=0.95\linewidth]{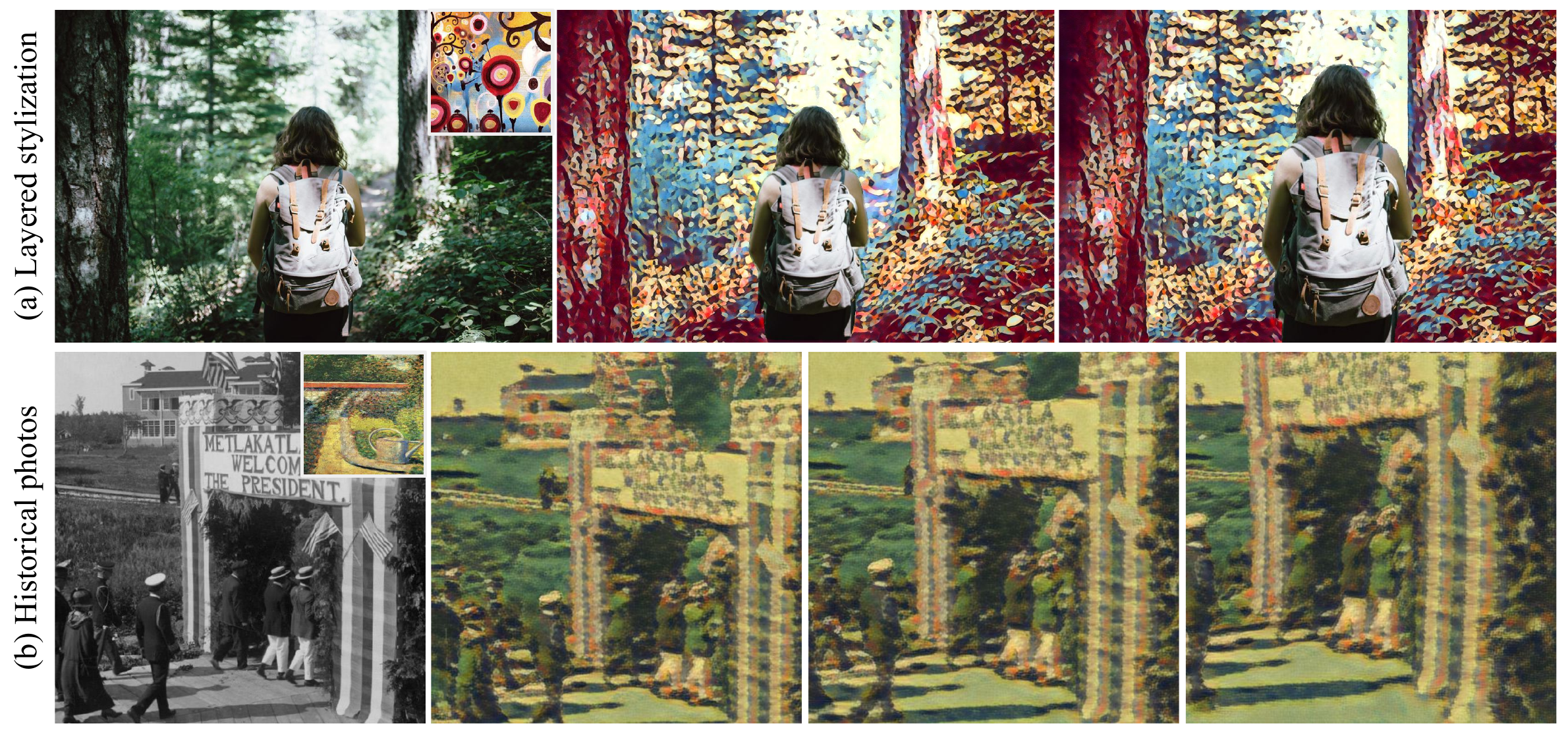}\vspace{-0.8em}
    \caption{{\bf Demonstration of Applications. } Layered stylization for AR {\it (upper)} and 3D browsing of a stylized historical photo\protect\footnotemark {\it (lower)}---``A small arch welcomes the President to Metlakatla, Alaska, created by D. L. Hollandy 1923.''}\vspace{-1.5em}
    \label{fig:application}
\end{figure}

\noindent {\bf Layered Stylization for AR applications. }
Human centered photography is of central interest in mobile AR applications. As a proof-of-concept experiment to demonstrate our method's potential in AR, we apply PointRend~\cite{kirillov2020pointrend} to segment foreground human subjects in images from Unsplash~\cite{unsplash2020}, and stylize the background scene using our method while leaving the foreground human untouched (Fig~\ref{fig:application}a). The final stylized 3D photo upon rendering initiates a virtual tour into a 3D environment in an artistic style.

\noindent {\bf 3D Exploration of Stylized Historical Photos. }
Historical photos represent a large fraction of existing image assets and remain under-explored in computer vision and graphics. As we demonstrate on the Keystone dataset~\cite{luo2020keystonedepth} (Fig~\ref{fig:application}b), our method can be readily applied for the 3D browsing of historical photos in an artistic style, bringing past moments back alive in an unexpected way.

\section{Discussion}
\label{sec:discuss}
In this paper, we connected neural style transfer and one-shot 3D photography for the first time, and introduced the novel task of 3D photo stylization -- generating stylized novel views from a single image given an arbitrary style. We showed that a na\"ive combination of solutions from the two worlds do not work well, and proposed a deep model that jointly models style transfer and view synthesis for high-quality 3D photo stylization. We demonstrated the strength of our approach using extensive qualitative and quantitative studies, and presented interesting applications of our method for 3D content creation. We hope our method will open an exciting avenue of applications in 3D content creation from 2D photos.
{\small
\bibliographystyle{ieee_fullname}
\bibliography{egbib}
}

\clearpage
\section*{Supplementary Material}

\setcounter{section}{0}
\renewcommand\thesection{\Alph{section}}

We refer to our {\it supplementary video} for an overview of our results and comparisons with the baselines. This document describes the technical details, the design of our user study, the details of extending our method to multi-view inputs, as well as a discussion on the limitation of our method.

\section{Implementation Details}

\begin{figure}
    \centering \vspace{-0.5em}
    \resizebox{1.0\columnwidth}{!}{
    \includegraphics{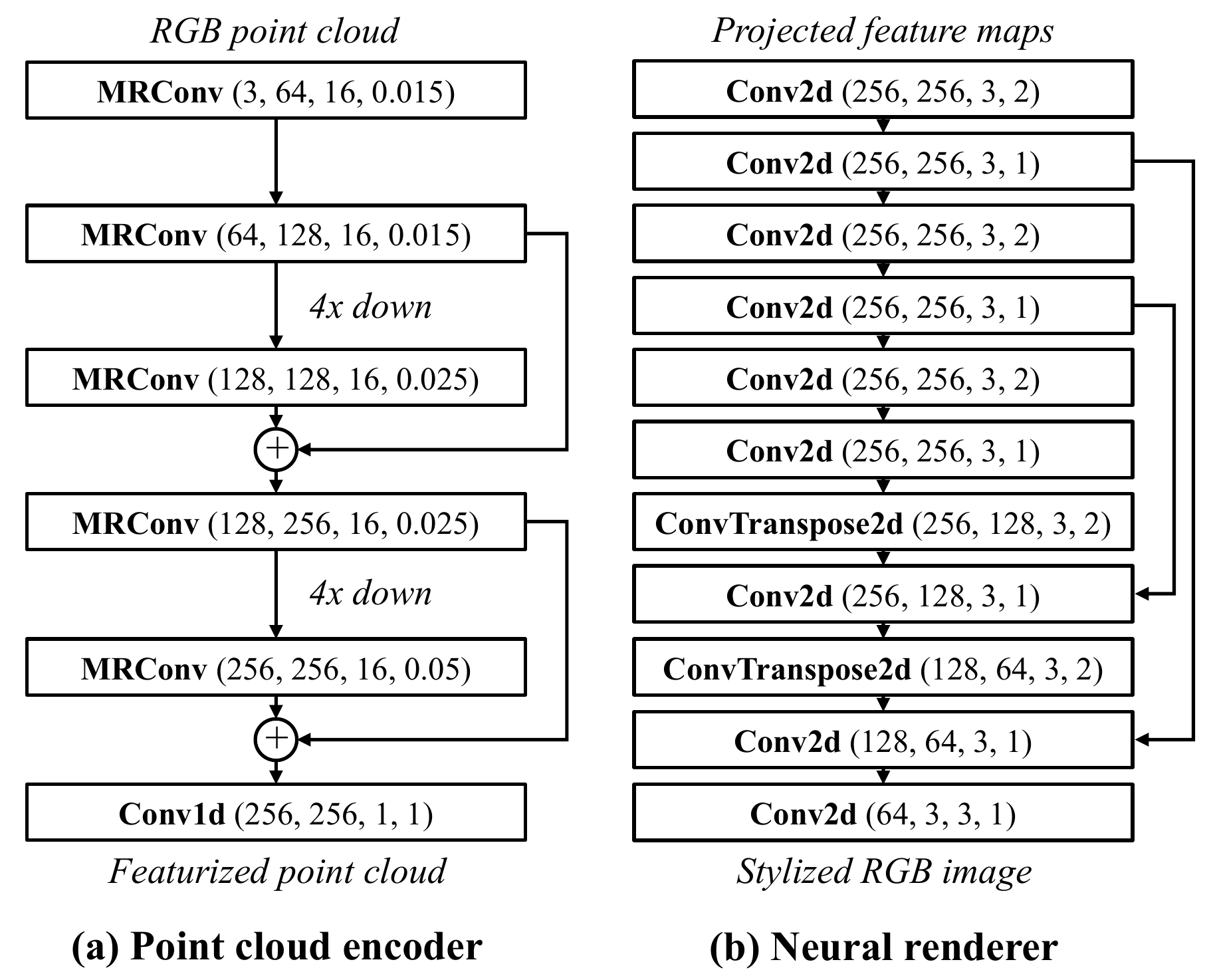}}\vspace{-0.5em}
    \caption{{\bf Model architecture. } Architecture of our point cloud encoder and neural renderer. The layer specifications are as follows: {\bf Conv1/2d} (input channel, output channel, kernel size, stride); {\bf MRConv} (input channel, output channel, maximum number of neighboring points, ball radius).}
    \label{fig:model}\vspace{-1.5em}
\end{figure}

Our model architecture is illustrated in Fig~\ref{fig:model}. We now present our implementation details.

\smallskip
\noindent {\bf Point Cloud Encoder Architecture. } Our GCN encoder adopts a hierarchical design for computational and memory efficiency. It takes an input RGB point cloud and processes it in three stages with 1, 2 and 2 MRConv layers~\cite{li2021deepgcns_pami} respectively. The point features are 64, 128 and 256 dimensional after each stage. Contrary to~\cite{li2021deepgcns_pami}, our MRConv variant performs point aggregation using ball queries, and we progressively increase the ball radius throughout the encoder to enlarge its receptive field. At the entry of each stage, we apply farthest point sampling to sub-sample the point cloud by a factor of 4. A residual connection is introduced every two layers to facilitate gradient flow during training. We apply batch normalization~\cite{ioffe15batchnorm} after each layer and use ReLU as the non-linearity.

\smallskip
\noindent {\bf Stylizer Architecture. } Our stylizer follows AdaAttN~\cite{liu2021adaattn}. As discussed in the main paper, we apply a multi-layer perceptron (MLP) with two fully-connected layers of 256 units to map content features to the style feature space before stylization. A symmetric MLP is applied after stylization to bring the modulated features back to the content feature space. The MLPs use ReLU as the non-linearity.

\smallskip
\noindent {\bf Neural Renderer Architecture. } Our neural renderer first up-samples the 256-dimensional encoder output via inverse distance weighted interpolation~\cite{qi2017pointnet++} until the output resolution is the same as the encoder input. The rasterizer~\cite{niklaus20193d} projects the up-sampled point features to the image plane of a novel view given camera pose and intrinsics. The resulting 2D feature maps have 256 dimensions and are further processed by a U-Net~\cite{ronneberger2015unet} with three levels. The encoder part of the U-Net downsamples the feature maps {\it without inflating the channel dimension}. We interpret this as a learnable anti-aliasing step in the same spirit as widely used super-sampling in computer graphics. The decoder part subsequently up-samples the feature maps via transposed convolution and meanwhile halves the channel dimension. The skip connections, implemented as $1\times1$ convs, pass along feature from the encoder to the decoder to facilitate gradient flow. All layers in the U-Net except the skip convs have a kernel size of $3\times3$. We apply leaky ReLU with a slope of 0.2 in the encoder and ReLU in the decoder as the non-linearity.

\begin{figure*}
    \centering
    \resizebox{0.9\textwidth}{!}{
    \includegraphics{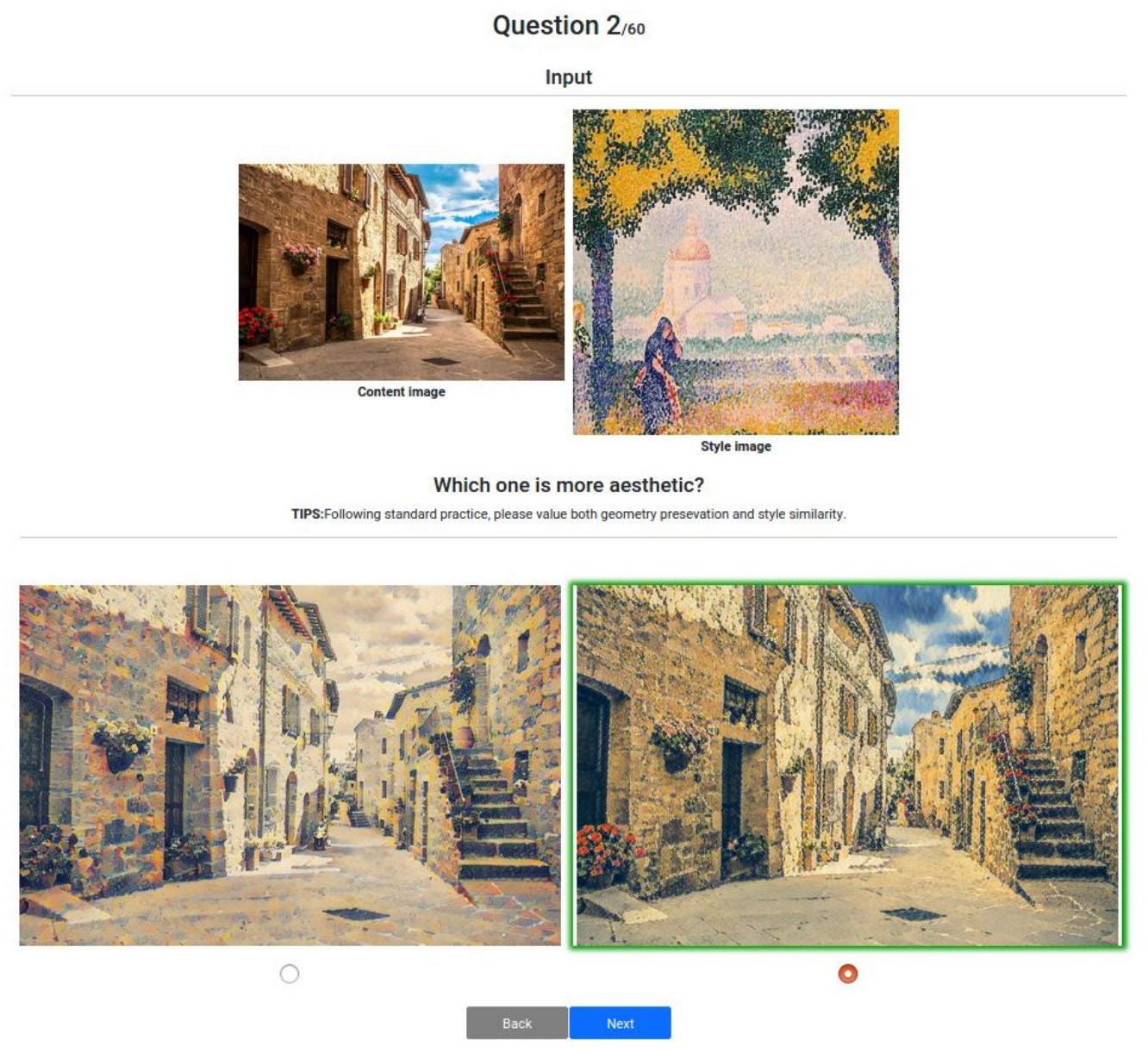}}\vspace{-1em}
    \caption{{\bf Screenshot of our user study. } A randomly picked question in our user study.}
    \label{fig:study}
\end{figure*}

\section{User Study Design}
We conduct a user study to compare our method with baselines that sequentially combine 3DPhoto~\cite{shih20203d} and one of the six image~\cite{huang2017arbitrary,li2019learning,liu2021adaattn} or video style transfer methods~\cite{wang2020consistent,deng2020arbitrary,liu2021adaattn}. The study includes three sections for the assessment of style quality, multi-view consistency and overall synthesis quality. Each section consists of 60 random binary choice questions that compare our method with one of the baselines. For convenience, a stylized 3D photo is displayed as a 90-frame snippet following a random camera trajectory. For fair evaluation of style quality, we only display stylized image of the input view so as not to bias participants toward more consistent renderings. Similarly, we hide the content and style images when consistency is evaluated to minimize the impact of style quality. Our analysis is based on a total of 5,400 votes collected from 30 volunteers. We show a screenshot of our user study in Fig~\ref{fig:study}. Our user study is anonymous and does not involve the collection of personally identifiable data.

\section{Details on Extension to Multi-view Inputs}
Extending our method to the multi-view setting is immediate after a small modification on point cloud normalization. Now that more than one input views are available, we back-project all views to a point cloud and transform it into the NDC space anchored to the center view. Everything else stays exactly the same, and importantly, the model need not be re-trained thanks to the normalization step. In our experiments, we use the same depth maps from~\cite{huang2021stylenvs} for a fair comparison with StyleScene \cite{huang2021stylenvs}. Those results were shown in Table 3 and Figure 10 of our main paper.

\section{Limitations}
Despite steady progress in monocular depth estimation, current state of the arts do not always produce reliable depth maps for complex scenes, and in particular for those pixels near depth discontinuities. Our method relies on monocular depth estimation on the input image and thus inherits the failure mode of the underlying depth estimators. As a partial remedy, we have demonstrated an extension of our method to use mutli-view inputs with more reliable depth estimations. Another limitation our method shares with StyleScene~\cite{huang2021stylenvs} lies in the run-time speed. While our method renders stylized images of 1K resolution at interactive rate on a TITAN Xp GPU, the current implementation may not support interactive exploration of a high-resolution stylized 3D photo on mobile devices. Future work may focus on improving rendering speed for 3D photo stylization.

\smallskip
\noindent {\bf Societal impacts}: We anticipate that our research would facilitate new applications of 3D content creation from 2D photos. Similar to other image manipulation methods like neural style transfer, our method might face potential copyright infringement, when copyright-protected content images are modified and improperly used.

\end{document}